\documentclass[10pt, a4paper]{article}
\usepackage{lrec2022} 
\usepackage{multibib}
\newcites{languageresource}{Language Resources}
\usepackage{graphicx}
\usepackage{tabularx}
\usepackage{soul}
\usepackage{titlesec}
\titleformat{\section}{\normalfont\large\bfseries\center}{\thesection.}{1em}{}
\titleformat{\subsection}{\normalfont\SmallTitleFont\bfseries\raggedright}{\thesubsection.}{1em}{}
\titleformat{\subsubsection}{\normalfont\normalsize\bfseries\raggedright}{\thesubsubsection.}{1em}{}
\renewcommand\thesection{\arabic{section}}
\renewcommand\thesubsection{\thesection.\arabic{subsection}}
\renewcommand\thesubsubsection{\thesubsection.\arabic{subsubsection}}

\usepackage{epstopdf}
\usepackage[utf8]{inputenc}

\usepackage{hyperref}
\usepackage{xstring}
\usepackage{ulem}
\usepackage{color}
\usepackage{xcolor}
\usepackage{arabtex}
\usepackage{multirow}
\usepackage{makecell}
\usepackage{utf8}
\usepackage[utf8]{inputenc}
\setcode{utf8}
\usepackage[export]{adjustbox}
\usepackage{color, colortbl}
\usepackage{booktabs}
\definecolor{LightCyan}{rgb}{0.95,1,1}
\definecolor{LightYellow}{rgb}{1,1,1}
\definecolor{LightGrey}{rgb}{1,1,1}
\usepackage{enumitem}
\usepackage{todonotes}

\usepackage{subcaption} 
\usepackage{fontawesome} 
\usepackage{comment}

\newcommand{\revfa}[1]{\textcolor{black}{#1}}

\title{ArCovidVac: Analyzing Arabic Tweets About COVID-19 Vaccination}

\name{Hamdy Mubarak$^1$, Sabit Hassan$^2$, Shammur Absar Chowdhury$^1$ and Firoj Alam$^1$} 

\address{$^1$Qatar Computing Research Institute, HBKU, Qatar \\
         $^2$University of Pittsburgh, USA \\
         \{hmubarak, shchowdhury, fialam\}@hbku.edu.qa, sah259@pitt.edu\\
         }

\abstract{
The emergence of the COVID-19 pandemic and the \textit{first global infodemic} have changed our lives in many different ways. We relied on social media to get the latest information about COVID-19 pandemic and at the same time to disseminate information. The content in social media consisted not only health related advise, plans, and informative news from policymakers, but also contains conspiracies and rumors. It became important to identify such information as soon as they are posted to make an actionable decision (e.g., debunking rumors, or taking certain measures for traveling). To address this challenge, we develop and publicly release the first largest manually annotated Arabic tweet dataset, \textit{ArCovidVac}, for the COVID-19 vaccination campaign, covering many countries in the Arab region. The dataset is enriched with different layers of annotation, including, \textit{(i)} Informativeness (more {\em vs.} less importance of the tweets); \textit{(ii)} fine-grained tweet content types (e.g., advice, rumors, restriction, authenticate news/information); and \textit{(iii)} stance towards vaccination (pro-vaccination, neutral, anti-vaccination). Further, we performed in-depth analysis of the data, exploring the popularity of different vaccines, trending hashtags, topics and presence of offensiveness in the tweets. We studied the data for individual types of tweets and temporal changes in stance towards vaccine. We benchmarked the ArCovidVac dataset using transformer architectures for informativeness, content types, and stance detection.
\\ \newline \Keywords{COVID-19, Vaccination, Stance Detection} }
\begin{document}

\maketitleabstract

\section{Introduction}
\label{sec:introduction}

Social media are integrated with our daily life. We share and access information through social media platforms making it the most prominent form of communication. Due to its reach to a larger and international population, many organizations and individuals use them to circulate their contents. Thus also making these platforms a constitutive part of online 
news distribution and consumption \cite{mitchell2014state}.

\begin{figure}[t]
\begin{center}
\includegraphics[scale=0.45,frame]{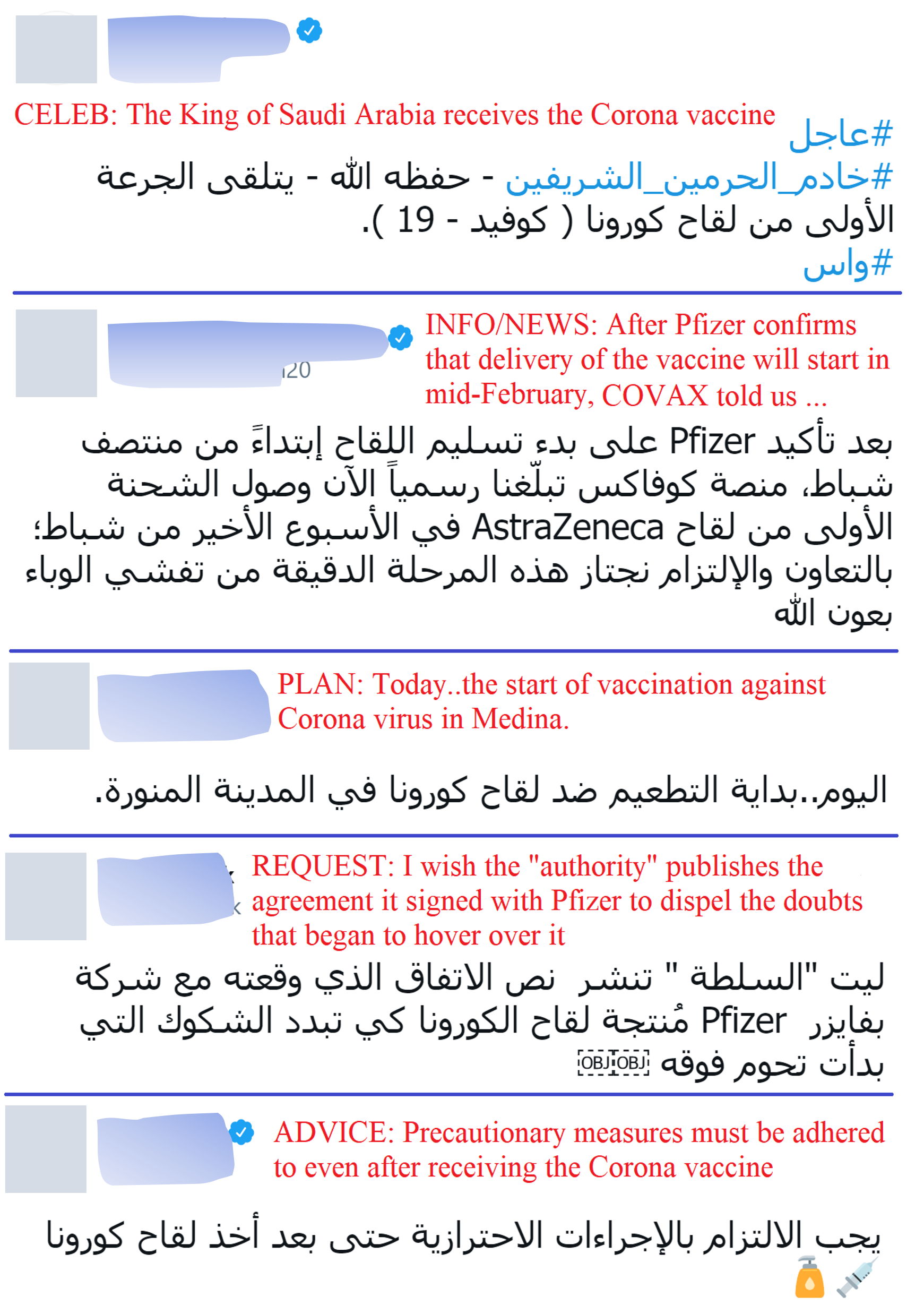} 
\caption{Examples for classes: Celebrity, Info-news, Plan, Requests and Advice}
\label{fig:class-examples1}
\end{center}
\end{figure}

\begin{figure}[t]
\begin{center}
\includegraphics[scale=0.45, frame]{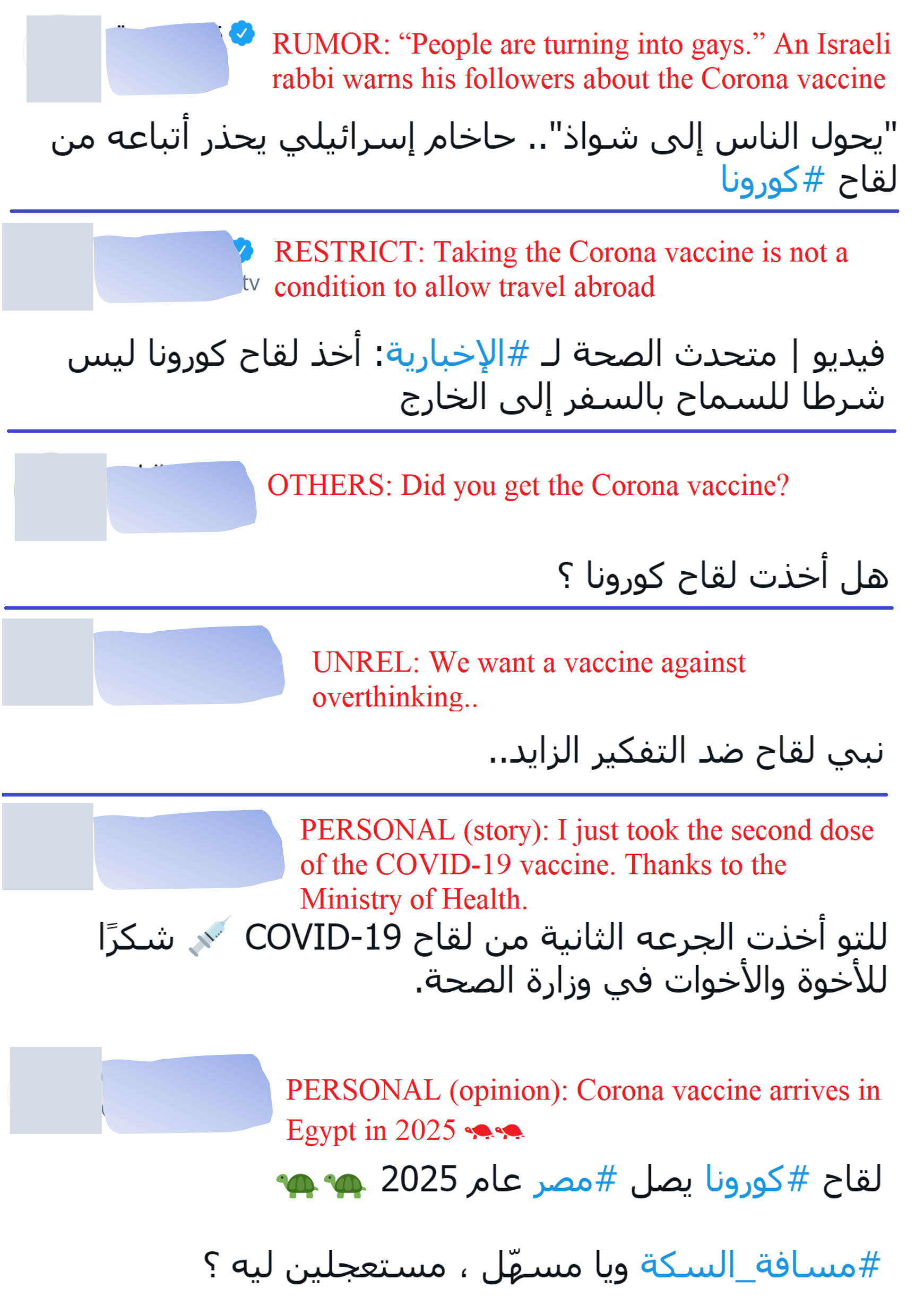} 
\caption{Examples for classes: Rumors, Requests, Others, Unrelated and Personal}
\label{fig:class-examples2}
\end{center}
\end{figure}

In Figure \ref{fig:class-examples1} and \ref{fig:class-examples2}, we demonstrate how online users share information (rumors, plan, travel restriction, personal experience). The post containing advice is important to reduce the spread as vaccinated people can be a carrier.  
Identifying such types of content from social media
become important to the government, international and local organisation for understanding psychological and physical well being along with public reactions to every taken actions.
Such an understanding can \textit{(i)} aid decision making by governments; and \textit{(ii)} prevent rumours and fake cures that can bring harm to the society.
Research studies have been conducted using numbers of COVID-19 datasets collected from Twitter. The research focused on: unlabeled~\cite{info:doi/10.2196/19273,Banda:2020,alqurashi2020large,haouari2020arcov19}, automatically labeled \cite{abdul2020mega,Umair2020geocovid19}, labeled using
distant supervision \cite{cinelli2020covid19,zhou2020repository}, and
small manually annotated \cite{song2020classification,vidgen2020detecting,shahi2020fakecovid,pulido2020covid,alam2020fighting,alam2021icwsmfighting} datasets.

Despite Arabic being one of the dominant languages on Twitter \cite{alshaabi2020a}, a very few research targeted toward aiding governments and international organisations in their decision making and understanding public perspective towards the vaccine, in the Arab region. 

To aid such decision making process, in this study, we designed and publicly released the largest manually annotated COVID-19 tweets regarding its vaccine and vaccination campaigns in the Arab region.
Our contributions can be summarized as follows: 
\begin{itemize}[noitemsep,topsep=0pt,leftmargin=15pt,labelwidth=!,labelsep=.5em]
\item We develop a large manually annotated COVID-19 vaccine infodemic, covering different countries in the Arab region, targeted to aid the policymakers and the society as a whole. To the best of our knowledge, this is the first dataset about COVID-19 vaccine in Arabic with diverse type of annotations.

\item We categorise the tweets for multiple classes (10 classes) including: plan, request, advice, restrictions, rumors, authenticate news or information, personal experience among others.         

\item We annotate the tweets, specifying their stance towards vaccine/vaccination process. We classify them into positive (pro-vaccination), negative (against vaccination) or as neutral stance.

\item We analyse the tweets for different annotated classes and explore what topics the content covers, top hashtags in each country, common sources that users post their tweets in different countries, etc. Moreover, we analyse the temporal changes in public stance on vaccination over time.

\item We benchmark the released dataset for several tasks. The classification tasks includes \textit{(i)} discriminating informative tweets from not-informative ones; \textit{(ii)} fine-grained multi-class tweet type categorisation; and \textit{(iii)} stance detection using transformer architectures.

\item We make our annotation guidelines, data and code freely available.\footnote{\url{https://alt.qcri.org/resources/ArCovidVac.zip}}
\end{itemize}


\section{Related Work}
\label{sec:related_work}
Research studies on COVID-19 focused on sentiment analysis \cite{yang2020senwave}, propagation of misinformation \cite{huang2020disinformation,shahi2020exploratory}, credibility check \cite{cinelli2020covid19,pulido2020covid,zhou2020repository}, detecting racial
prejudices and fear \cite{Medford2020.04.03.20052936,vidgen2020detecting} along with
situational information, e.g., caution and advice
\cite{li2020characterizing}. Moreover, studies also include detecting mentions and
stance with respect to known misconceptions
\cite{hossain-etal-2020-covidlies}. 

These studies relies mostly on the social media dataset -- mainly Twitter using queries or some distant supervision. Most of the large-scale COVID-19 datasets are unlabeled tweet collection, including multi-lingual dataset of 123M tweets \cite{info:doi/10.2196/19273}, 152M tweets \cite{banda2020largescale}, a billion multilingual tweets \cite{abdul2020mega} and GeoCoV19 \cite{Qazi_2020} containing 524M tweets with their location information. 
In addition to unlabeled data, some dataset are created using
distant supervision \cite{cinelli2020covid19,zhou2020repository} and some manually annotated \cite{song2020classification,vidgen2020detecting,shahi2020fakecovid,pulido2020covid}.

For Arabic, we see a similar trend in developing datasets. 
The Arabic dataset, studied in \cite{alqurashi2020large} provide a large dataset of Arabic tweets containing keywords related to COVID-19. Similarly, ArCOV-19 proposed in \cite{haouari2020arcov19}, contains 750K tweets obtained by querying Twitter. The manually labeled datasets are relatively fewer and also diversity of annotated labels is little to none. 
Authors in \cite{alam2020fighting,alam2021icwsmfighting} manually annotated tweets in multiple languages for fact-checking, harmfulness to society, and the relevance of the tweets to governments or policy makers. Another study in \cite{alsudias-rayson-2020-covid} collected 1M unique Arabic tweets related to COVID-19 in the early 2020, among which a random 2000 tweets are annotated for rumor detection based on the tweets posted by the Ministry of Health in Saudi Arabia. Authors in \cite{mubarak2020arcorona} annotated 8K tweets and labeled them for different types of content.
In \cite{yang2020senwave}, author also annotated 10K Arabic and English tweets for the task of fine-grained sentiment analysis. 

\paragraph{Our Dataset:} Prior studies are mainly focused on one or two aspects of actionable information (e.g., factuality, rumor detection). In comparison, our work is focused on \textit{Informativeness}, \textit{fine-grained content types} and their \textit{stance towards vaccination} with manual annotation of ($10K$) Arabic tweets. Such a diversity of labels enables the community to design and develop models in a multitask learning setup (i.e., fine-grained content types and stance in one model).

In addition, we specifically developed the dataset targeting the vaccination campaigns in the Arab region, covering many countries. Unlike \cite{mubarak2020arcorona}, we annotated the stance of the tweets. We also manually analysed different annotated classes, explored topics and temporal changes in stance regarding vaccines. We also studied the classification errors for the stance and tweet type classification.



\section{Data Collection}
\label{sec:data_collection}
Fighting the pandemic as well as infodemic requires to identify and understand the content shared on social media either to reduce the spread of harmful content, health related disinformation or to make an actionable decision (attention worthy content for policymakers) \cite{alam2020fighting,alam2021icwsmfighting}. Such an understanding can help in identifying concerns and rumors about vaccination, public sentiment, requests from governments and health organizations, etc, while facilitating policymakers.
This is a challenging given that manually annotated language specific (e.g., Arabic) datasets are scares. 



To address this challenge, 
we collected Arabic tweets and manually annotated them. 
To collect the tweets we used the following keywords: \<تطعيم، لقاح، مطعوم> (vaccine, vaccination) between \textit{Jan 5\textsuperscript{th}} and \textit{Feb 3\textsuperscript{rd} 2021}.\footnote{Words used in different countries in the Arab World.} We used twarc search  API\footnote{https://github.com/DocNow/twarc} to collect these tweets specifying Arabic language. Our data collection timeline coincides with the phase where many Arab countries already started their COVID-19 vaccination campaigns.\footnote{\url{https://tinyurl.com/mtm4wtrh}} For example, Saudi Arabic (SA)\footnote{ISO 3166-1 alpha-2 for country codes: \url{https://tinyurl.com/mubpbjx6}} started vaccine rollout in the middle of Dec 2020.




We collected 550K unique tweets in total. After considering only tweets that were liked or retweeted at least 10 times, we ended up with 14K tweets. We assume that tweets with large number of likes or retweets are the most important ones as they get highest attention from Twitter users. Out of them, 10K tweets were randomly chosen for manual annotation.

\section{Data Annotation}
\label{sec:data_annotation}

\subsection{Annotation Task and Labels}
\revfa{
We manually analyzed the random samples of selected tweets to understand the data at hand and to design and define the annotation task and class labels. Note that, we identified different types of class labels based on our engagement with the ministry of public health and policymakers.
For the three types of labels, we manually annotate two types \textit{fine-grained content types} and their \textit{stance}, and the informativeness type labels are inferred from fine-grained content types. Below, we define the class labels with examples, which are given to the annotators as instructions. We asked the annotators to follow these definitions while annotating the tweets. 
}

\paragraph{Fine-grained Content Types:}
\begin{enumerate}[noitemsep,topsep=0pt,leftmargin=15pt,labelwidth=!,labelsep=.5em]
    \item \textbf{Info-news:} Information and news about vaccine and conditions of taking 
    \item \textbf{Celebrity:} Vaccination of celebrities such as politicians, artists,  and public figures
    \item \textbf{Plan:} Governments’ vaccination plans, vaccination progress and reports
    \item \textbf{Requests:} Requests from governments, ex: speedup vaccination process
    \item \textbf{Rumors:} Rumors and refute rumors
    \item \textbf{Advice} Advice or instructions related to the virus or its vaccination
    \item \textbf{Restrictions:} Restrictions and issues that will be affected by taking vaccine, ex: travel
    \item \textbf{Personal:} Personal story or opinion about the vaccine, ex: thank government
    \item \textbf{Unrelated:} Unrelated to vaccination process. This includes also spam and ads
    \item \textbf{Others} Related to vaccine but not listed in the above classes
\end{enumerate}

\paragraph{Informativeness:}
For informativeness, the former seven class labels are considered as \textit{more informative} and the later three class labels are considered as \textit{less informative}.

\paragraph{Stance:}
For identifying stance we use the following labels: 

\begin{itemize}[noitemsep,topsep=0pt,leftmargin=15pt,labelwidth=!,labelsep=.5em]
    \item \textbf{Positive:} Support vaccination, encourage people to take vaccine, and remove their fears.\\
    \textit{Example:} \<متحدث الصحة: المشككون في فعالية لقاح >\\
    \<كورونا سوف يأتون لأخذ اللقاح> \\
    \textcolor{blue}{\textit{Health spokesperson: Those who doubt the effectiveness of the Corona vaccine will come and get it}}
    \item \textbf{Negative:} Oppose vaccination and fear people from vaccine\\
    \textit{Example:} \<قلق بالغ في النرويج بسبب وفاة > \\
    \<23 شخصا بعد تلقيهم لقاح فايزر> \\
    \textcolor{blue}{\textit{Extreme concern in Norway because 23 people have died after receiving the Pfizer vaccine}}
    \item \textbf{Neutral/Unclear:} Neither clearly support nor oppose vaccination\\
    \textit{Example:} \<توتر العلاقات بعد رفض بريطانيا >\\
    \<تسليم فرنسا 15 مليون من لقاح كورونا> \\
    \textcolor{blue}{\textit{Relations are strained after Britain refused to deliver 15 million doses of the Corona vaccine to France}}
\end{itemize}

\subsection{Manual Annotation}
\revfa{
For the manual annotation, we opted to use Appen crowdsourcing platform\footnote{\url{www.appen.com}}. One of the challenges with crowdsourced annotation is to find a large number of qualified workers while filtering out low-quality workers or spammers~\cite{chowdhury2015selection,chowdhury2014cross}. To deal with this problem and to ensure the quality of the annotation we followed standard evaluation~\cite{chowdhury2020multi}, i.e., we used 150 gold standard test tweets. Based on these gold standard test tweets, each annotator needed to pass at least 70\% of the tweets to participate in the annotation task. Given that the content of the tweet is in Arabic, therefore, we only allowed Arabic speaking participants from all Arab countries. While annotators needed pass such criteria to annotate each tweet, we also designed the annotation task to label each tweet by three annotators so that final label can be selected based on the majority agreement. 
}






In total, 245 annotators participated in the annotation task from different Arab countries.\footnote{We paid more than \$15 per hour of work to conform to the minimum wage rate in US.} 

\revfa{
We selected the final label for each tweet based on the label agreement score\footnote{\url{https://success.appen.com/hc/en-us/articles/360038386492-How-to-Calculate-Overall-Unit-Agreement}} greater than or equal to 70\%. In Table \ref{tab:classes}, we report the distribution of the dataset. As mentioned earlier, the class labels for \textit{Informative} are inferred from fine-grained labels. 
}

\paragraph{Annotation agreement:} We compute the annotation agreement using Cohen’s kappa coefficient, and found an agreement score of 0.82, 
which indicates high annotation quality.




\begin{table}[]
\centering
\scalebox{0.95}{
\begin{tabular}{@{}lrlr@{}}
\toprule
\multicolumn{1}{c}{\textbf{Class}} & \multicolumn{1}{c}{\textbf{Count}} & \multicolumn{1}{c}{\textbf{Class}} & \multicolumn{1}{c}{\textbf{Count}} \\ \midrule
\multicolumn{2}{c}{\textbf{Fine-grained}} & \multicolumn{2}{c}{\textbf{Informative}} \\ \midrule
Info-news & 5,225 & More Informative & 7,891 \\
Celebrity & 1,398 & Less Informative & 2,109 \\
Plan & 860 & \textbf{Total} & 10,000 \\ \cmidrule{3-4}
Requests & 172 & \multicolumn{2}{c}{\textbf{Stance}} \\\cmidrule{3-4}
Rumors & 118 & {Positive} & 7,968 \\
Advice & 94 & {Negative} & 638 \\
Restrictions & 24 & {Neutral/Unclear} & 1,396 \\
Personal & 1,430 & \textbf{Total} & 10,002 \\\cmidrule{3-4}
Unrelated & 450 &  &  \\
Others & 229 &  &  \\
\textbf{Total} & 10,000 &  &  \\ \bottomrule
\end{tabular}
}
\caption{Distribution of the annotated class labels. 
}
\label{tab:classes}
\end{table}

\section{Analysis}
\label{sec:analysis}

%
%
%


We present in-depth analysis of the ArCovidVac dataset, highlighting the vaccine popularity, the most common trending hashtags in the dataset for different countries. Moreover, we studied the popularity of mobile applications used to fight spreading of the virus.
We also analyze common rumors and requests from governments followed by topic and country distribution of tweets and the most common sources in each country. To our believe, this analysis gives a broad understanding how people are reacting 
the vaccine campaigns in each individual countries.

\paragraph{Vaccine Popularity:} Table \ref{tab:vaccine-hastags} shows the list of vaccine hashtags mentioned in our dataset. This can give a rough estimate about vaccine popularity in the Arab countries during the period of our study. More details about these vaccines can be found at: \url{https://en.wikipedia.org/wiki/COVID-19_vaccine}\\

\paragraph{Trending Hashtags:} The most frequent hashtags in different countries are listed in Table \ref{tab:hashtags}. The main messages in these hashtags show worry from vaccination, advice to take precautionary measures, and reassure people that vaccine is safe.

\begin{table}
\centering
\scalebox{0.65}{
\begin{tabular}{lrrc}
\hline
\textbf{Vaccine} & \textbf{Top Hashtags} &  \textbf{\#} & {\textbf{CC}}\\ \hline
\textbf{Pfizer} & Pfizer, \<فايزر، لقاح\_فايزر، بيونتيك، باينوتيك> & 184& US\\

\rowcolor{LightYellow}
\textbf{AstraZeneca} & \<استرازينيكا، اوكسفورد، استرازينكا>  & 94& UK\\

\textbf{Sputnik V} & 5\<سبوتنيك، سبوتنيك> & 65&RU\\

\rowcolor{LightYellow}
\textbf{Moderna} & Moderna, \<موديرنا، مودرنا> & 43&US\\

\textbf{BBIBP-CorV} & Sinopharm, \<سينوفارم> & 24&CN\\

\rowcolor{LightYellow}
\textbf{CoronaVac (Sinovac)} & \<سينوفاك، كورونافاك> & 10& CN\\

\textbf{Johnson \& Johnson} & \<جونسون، جونسون\_اند\_جونسون> & 5&US\\

\rowcolor{LightYellow}
\textbf{Novavax} & \<نوفافاكس> & 2&US\\
\hline
\end{tabular}
}
\caption{Vaccine hashtag frequencies. Arabic hashtags are mainly  different transliterations of vaccine names. CC: Country Code of the manufacturing company.}
\label{tab:vaccine-hastags}
\end{table}

\begin{table*}
\centering
\scalebox{0.80}{
\begin{tabular}{lrlr}
\hline
\textbf{Country} & \textbf{Hashtags} & \textbf{Translation} & \textbf{\#}\\ \hline
\textbf{IQ} & \<نريد\_لقاح\_آمن> & We want a safe vaccine & 288 \\

\rowcolor{LightYellow}
\textbf{SA} & \<الملك\_يتلقي\_لقاح\_كورونا، نعود\_بحذر> & The king takes COVID vaccine, We return cautiously & 174 \\

\textbf{LB} & \<لقاح\_آمن، خليك\_بالبيت> & Safe vaccine, Stay home & 157 \\

\rowcolor{LightYellow}
\textbf{AE} & \<يدا\_بيد\_نتعافى، اخترت\_التطعيم> & Hand in hand we recover, I chose vaccination & 151\\

\textbf{EG} & \<معا\_نطمئن> & Together we can rest assured & 7\\

\rowcolor{LightYellow}
\textbf{MA}& \<نبقاو\_على\_بال> & We remain alert & 7 \\

\textbf{OM} & \<عمان\_تواجه\_كورونا، التحصين\_وقاية> & Oman fights Corona, Vaccination is protection & 6 \\

\rowcolor{LightYellow}
\textbf{JO} & \<المطعوم\_وقاية، صحتك\_بتهمنا> & Vaccine is protection, Your health is important to us & 5\\
\hline
\end{tabular}}
\caption{Most frequent hashtags in some Arab countries.}
\label{tab:hashtags}
\end{table*}

\begin{figure}[!h]
\begin{center}
\includegraphics[scale=0.40, frame]{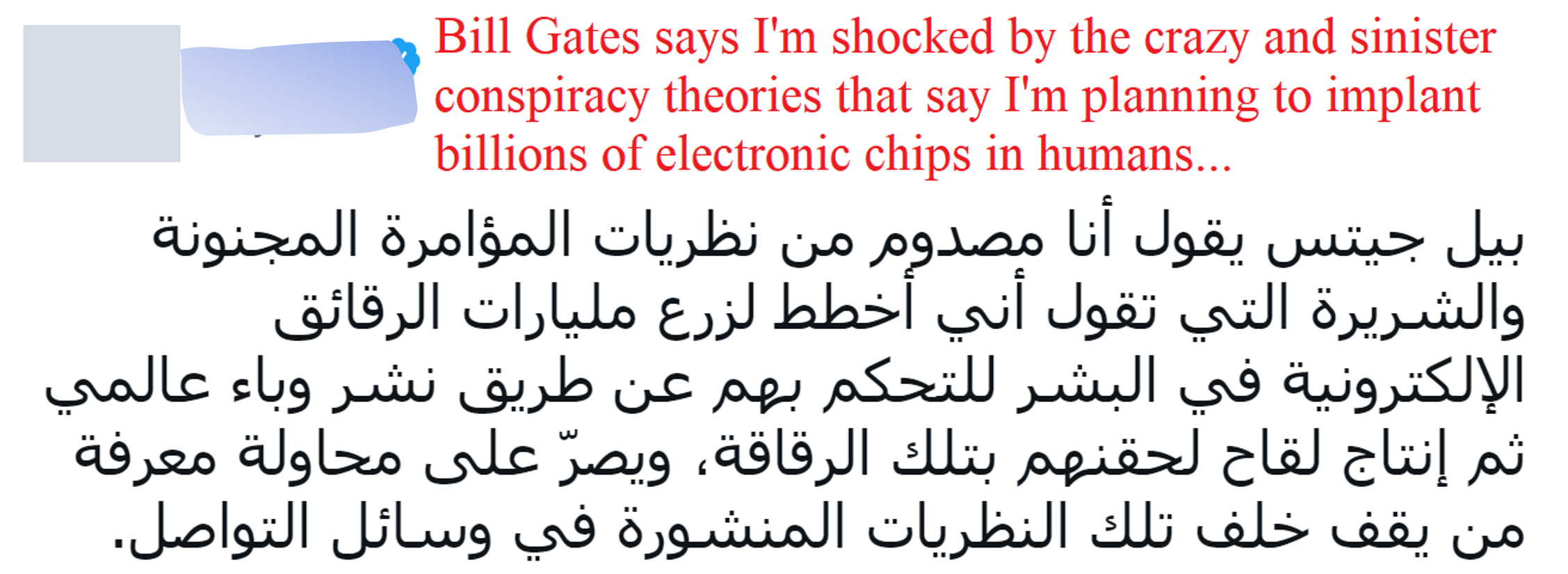} 
\caption{An example of a tweet with \textit{conspiracy theory}.}
\label{fig:rumor}
\end{center}
\end{figure}

\paragraph{Rumors:}
Rumors are very important class that needs more attention from governments and policy makers. False claims about vaccines can negatively affect public trust in vaccination campaigns. This may cause a threat to global public health. We analyzed all rumors in our dataset and classified them into the following main topics:

\begin{itemize}[noitemsep,topsep=0pt,leftmargin=15pt,labelwidth=!,labelsep=.5em]
    \item \textbf{Vaccine is unsafe and ineffective:} i) causes death and has side effects especially on elderly; ii) manipulates genes; iii) causes infertility in women.
    
    \item \textbf{Conspiracy theory:} i) big countries or companies created the virus and its vaccine for commercial purposes; ii) vaccine has chips to monitor and control  people; iii) vaccine is a biological weapon; 
    iv) question about finding vaccines within a year. Figure \ref{fig:rumor} shows the most retweeted and targeted tweet in this category.
    
    \item \textbf{Doubts} about government statistics,  plans, and vaccination process.
\end{itemize}

\begin{figure}[!h]
\begin{center}
\includegraphics[scale=0.40,frame]{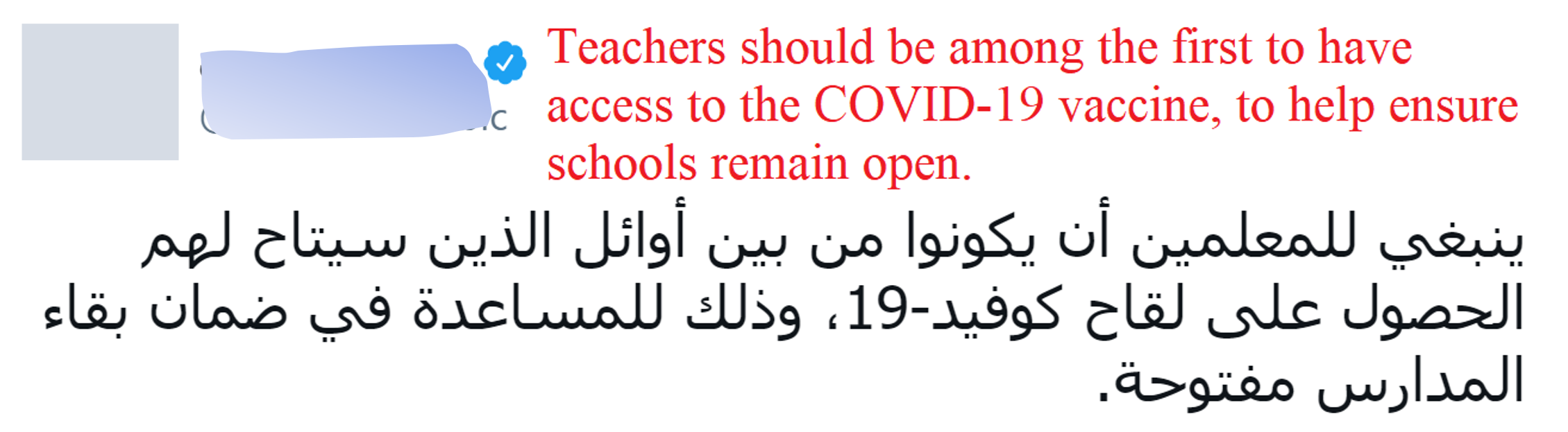} 
\caption{An example of a tweet with a request to give priority to the teaching professional.}
\label{fig:priority}
\end{center}
\vspace{-0.4em}
\end{figure}


\paragraph{Requests from Governments:}
We analyzed all requests from governments and classified their main topics into the following classes:

\begin{itemize}[noitemsep,topsep=0pt,leftmargin=15pt,labelwidth=!,labelsep=.5em]
    \item \textbf{Safe vaccine:} i) wait until studies and other countries prove vaccine effectiveness and safety; ii) prefer US vaccines over their  Chinese counterparts; iii) refuse vaccine from the US (especially in Iraq).
    
    \item \textbf{Fair access to vaccine:} i) rich and poor countries and people; ii) males and females; iii) citizens, expats and refugees; iv) cities and regions in the same country; v) politicians and common people; vi) Israel and Palestinians.
    
    \item \textbf{Vaccination process:} i) speedup; ii) transparency in plans and contract details; iii) finding alternative companies and cheaper vaccines; iv) allow private sector to sell vaccines.
    
    \item \textbf{Give priority} to some professionals 
    such as doctors, teachers, players, and  natives. Figure \ref{fig:priority} shows one of the most common tweets that asks to give priority to the teaching professional. 
\end{itemize}

\begin{figure}[!h]
\begin{center}
\includegraphics[scale=0.40, frame]{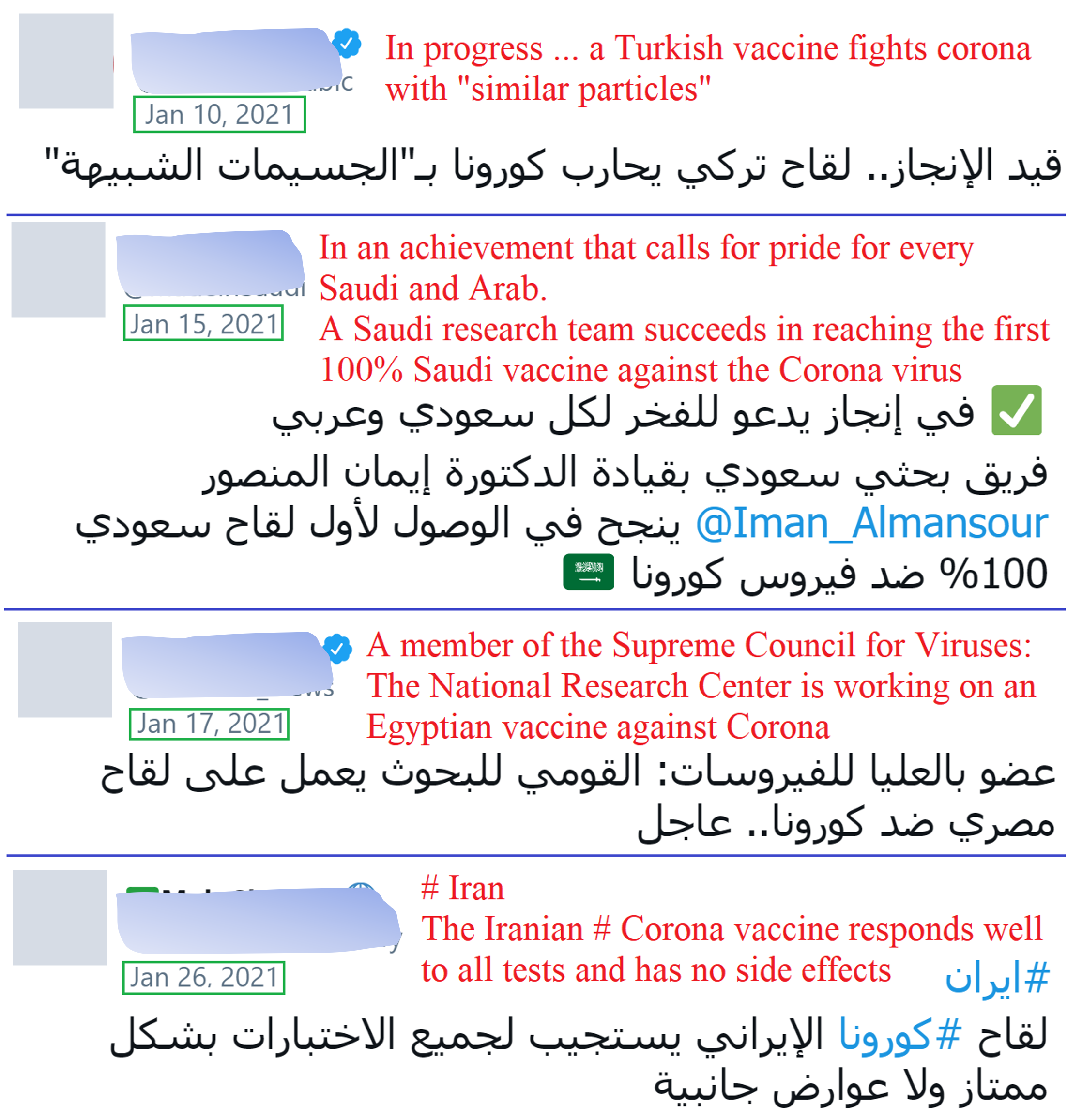} 
\caption{Vaccine announcements in Turkey, Egypt, Saudi Arabia and Iran.}
\label{fig:vaccine-announcements}
\end{center}
\end{figure}

\paragraph{Vaccine Announcements:}
We spotted many news, posted in Jan 2020, about successful vaccines coming from research labs in different countries in the MENA region, but in reality none of those vaccines was used in any Arab countries until the date of our study. Examples of such announcements are shown in Figure \ref{fig:vaccine-announcements}. We suspect these news were posted for political or social purposes.

\begin{figure}[!h]
\begin{center}
\includegraphics[scale=0.36]{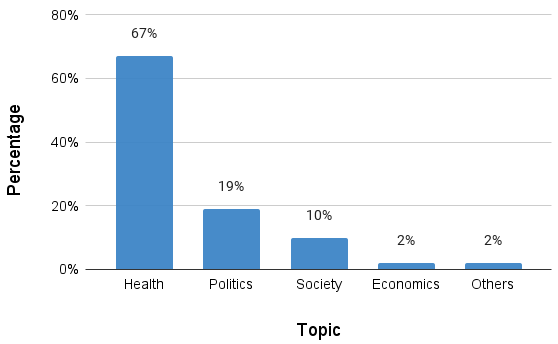} 
\caption{Distribution of \textit{topics} labeled on a random sample of 1000 tweets.}
\label{fig:topics}
\end{center}
\end{figure}

\paragraph{Topic and Country Distribution:}
\revfa{We took a random sample of 1000 tweets and manually categorized them by expert annotators for their main topics, such as health, politics, society and economy.}
Additionally for all tweets, we use ASAD \cite{hassan-etal-2021-asad}, which achieves 88.1\% F1 score on the UL2C dataset \cite{mubarak-hassan-2021-ul2c} for country prediction of original tweet authors based on their user locations in their profiles.
Figure \ref{fig:topics} shows that in addition to the health topics in most of the tweets, one third of tweets talk about the vaccine from different aspects (e.g., attacking politicians or countries). We also found that 7\% of tweets have hate speech or offensive language. 

Country distribution of tweets and top accounts that users share their posts the most in each country are shown in Table \ref{tab:country}. 
Analysis of such accounts shows that people retweet posts mainly from online news agencies and newspapers in their countries, and less from some journalists or activists. Most of those accounts are verified. Surprisingly, accounts of ministries of health were not among the top four sources in the listed countries. We anticipate one reason for that might be due to the less amount of posts from ministries of health compared to the large volumes of tweets that come from news agencies and newspapers.

\begin{table}
\centering
\scalebox{0.75}{
\begin{tabular}{@{}lrl@{}}
\toprule
\textbf{CC} & \textbf{\%} &  \textbf{Top Accounts}\\ \midrule
\textbf{SA} & 25  & sabqorg, Akhbaar24, KSA24, ajlnews
\\
\rowcolor{LightYellow}
\textbf{AE} & 14  & cnnarabic, AlArabiya\_Brk, skynewsarabia, AlHadath
\\

\textbf{LB} & 11  & \begin{tabular}[c]{@{}l@{}}AlMayadeenNews,	ALJADEEDNEWS,	JamalCheaib \end{tabular}\\ 

\rowcolor{LightYellow}
\textbf{EG} & 8  & youm7, 	AlMasryAlYoum, RassdNewsN, Extranewstv
\\

\textbf{GB} & 5  & aawsat\_News, AlarabyTV, 	IndyArabia, Mhd\_AlObaidi\\

\rowcolor{LightYellow}
\textbf{KW} & 5  & liferdefempire, WhistleBlowerQ8, gucciya234, TfTeeeSH
\\
\textbf{JO} & 4  & AlMamlakaTV, alrai, khaberni, RoyaTV\\

\rowcolor{LightYellow}
\textbf{TR} & 4  & TRTArabi, aa\_arabic, TurkPressMedia, YeniSafakArabic\\

\textbf{DZ} & 3  & ennaharonline, El\_Bilade, 	radioalgerie\_ar, elkhabarlive\\
\rowcolor{LightYellow}
\textbf{RU} & 3  & RTarabic, RTarabic\_Bn \\
\bottomrule
\end{tabular}
}
\caption{Distribution of top accounts across different countries. CC: Country Code.}
\label{tab:country}
\end{table}

\begin{figure}[!h]
\begin{center}
\includegraphics[scale=0.15,]{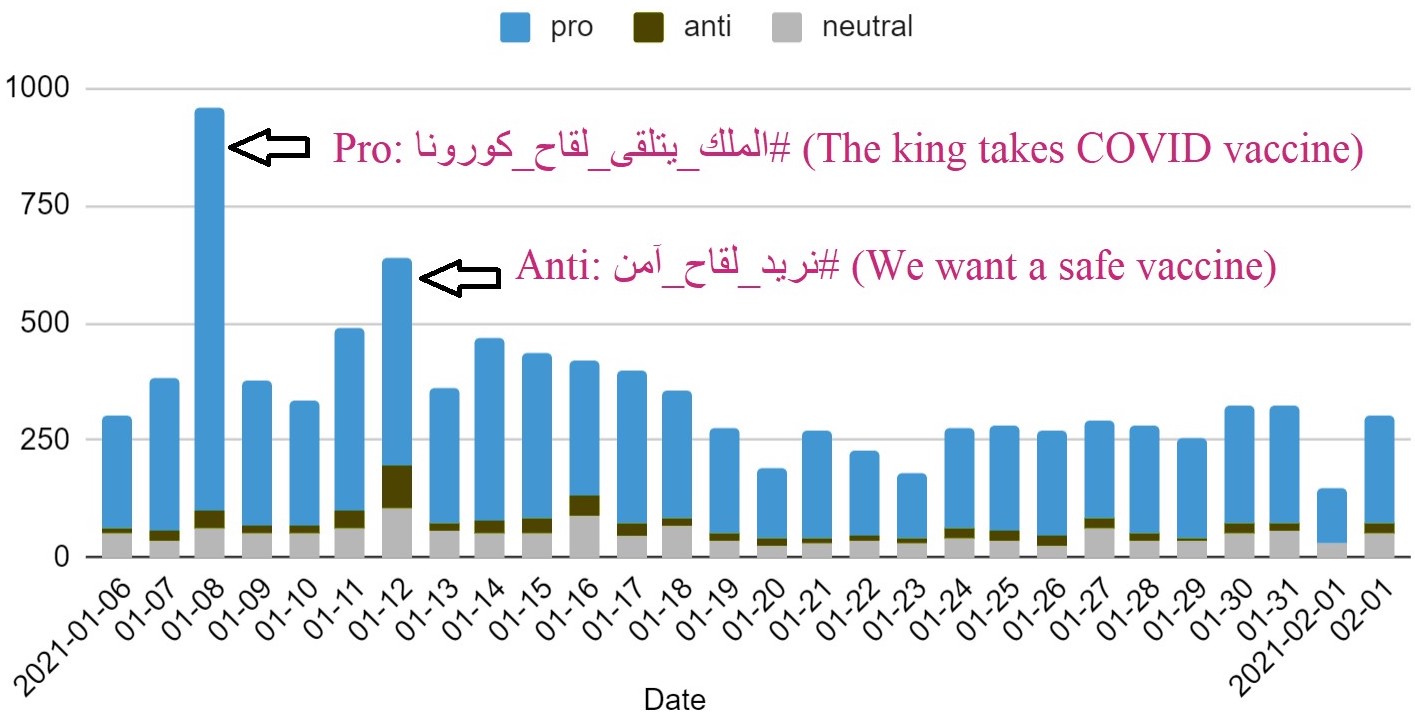} 
\caption{Distribution of \textit{stance} towards vaccine over time. pro: positive stance, anti: negative stance.}
\label{fig:stance-timeline}
\end{center}
\end{figure}

\paragraph{Distribution of Stance:}
Figure \ref{fig:stance-timeline} shows timeline of stance towards vaccine during the period of our study. We observe a big increase of positive stance (pro vaccine) in Jan 8\textsuperscript{th} when media announced that the king of Saudi Arabia took the vaccine. This can show the effect of sharing news about celebrity vaccination on public opinion. On the opposite side, we found an increase of negative stance (anti vaccine) in Jan 12\textsuperscript{th} than other days due to wide adoption of a hashatg against US vaccines among activists especially in Iraq.

\paragraph{Uses of Mobile Application:} We spotted also discussions about the mobile applications listed in Table \ref{tab:applications} that help fighting the spread of COVID-19 virus. The purposes of these applications vary from showing the health status of application users, reporting violations of precautionary measures, booking and following-up medical services, tracking medicines, and facilitating travel/visa process. It is worth to mention that there are other applications used in different Arab countries but didn't appear in our dataset, e.g., \<احتراز> Ehterhaz ``Precaution'' (QA,  released in Apr'20), \<أمان> Aman ``Safety'' (JO, Aug'20), etc. 

\begin{table} [!htb]
\centering
\scalebox{0.65}{
\begin{tabular}{lrllr|r}
\toprule
\textbf{Application (and meaning)} & \textbf{Arabic Name} & \textbf{CC} & \textbf{Date} & \textbf{\#} & \textbf{DL}\\ \midrule
\textbf{Tawakkalna} & \<توكلنا> & SA & May'20& 35 & 10M\\
(We Trust in God) & & &\\
\textbf{Sehhaty}  & \<صحتي> & SA & Dec'20 & 32 & 5M\\
(My Health) & & &\\

\textbf{Kuwait Mosafer} & \<كويت مسافر> & KW & Feb'21 &  3 & 5K\\
(Kuwait Traveller) & & &\\
\textbf{DHA} & \<صحة دبي> & AE & Dec'20& 2 & 500K\\
(Dubai Health Authority) & & & \\

\textbf{Al Hosn UAE}  & \< الحصن> & AE & Apr'20& 1 & 1M \\
(The Fort) & & &\\
\bottomrule
\end{tabular}
}
\caption{Applications used to fight COVID-19 in some Arab countries. DL: Downloads at Google Store in May'20. 
}
\label{tab:applications}
\end{table}

\section{Experiments}


\begin{table}[]
\centering
\scalebox{0.85}{
\begin{tabular}{lrrr}
\toprule
\multicolumn{1}{c}{\textbf{Class}} & \multicolumn{1}{c}{\textbf{Train}} & \multicolumn{1}{c}{\textbf{Dev}} & \multicolumn{1}{c}{\textbf{Test}} \\\midrule
\multicolumn{4}{c}{\textbf{Informativeness}} \\ \midrule
More informative & 5482 & 819 & 1590 \\
Less informative & 1518 & 181 & 410 \\ \midrule
\multicolumn{4}{c}{\textbf{Fine-grained categorization}} \\ \midrule
Info-news & 3623 & 545 & 1057 \\
Celebrity & 977 & 145 & 276 \\
Plan & 606 & 82 & 172 \\
Requests & 112 & 20 & 40 \\
Rumors & 79 & 15 & 24 \\
Advice & 67 & 10 & 17 \\
Restrictions & 18 & 2 & 4 \\
Personal & 1027 & 128 & 275 \\
Unrelated & 324 & 36 & 90 \\
Others & 167 & 17 & 45 \\ \midrule
\multicolumn{4}{c}{\textbf{Stance}} \\ \midrule
Negative & 439 & 70 & 127 \\
Neutral & 1017 & 126 & 253 \\
Positive & 5544 & 804 & 1620 \\
\bottomrule
\end{tabular}
}
\caption{Distribution of labels for different tasks.}
\label{tab:distribution}
\end{table}

\revfa{
For the experiments, we randomly split the data into train, dev and test sets with 7000, 1000 and 2000 tweets, respectively. Table \ref{tab:distribution} shows the distribution of labels across the three label sets, defined as three tasks, which include
{\em(i)} \textit{Task1:} distinguish important tweets from less important ones, {\em(ii)} \textit{Task2:} fine-grained classification of important tweets, and {\em(iii)} \textit{Task3} stance of the tweets.
}

\revfa{
We train several models, SVM with different features combinations, and different tranformers models as discussed below. 
}

To measure the performance of the models we compute and report macro-averaged Precision (P), Recall (R) and F1 score along with Accuracy (Acc) on test set.
We use F1 score as the primary metric for comparison.

\label{sec:experiments_results}
\subsection{Classification Models}
\paragraph{Support Vector Machines (SVMs)} SVMs are known to perform decently for Arabic text classification tasks, with imbalanced class distribution, in tasks such as offensiveness detection \cite{hassan-etal-2020-alt-semeval,chowdhury2020multi}, text categorisation \cite{chowdhury2020improving} or dialect identification \cite{abdelali2020arabic}. Due to its popularity and efficacy among machine learning algorithms, this is one of the algorithm we explored in this study for the aforementioned classification tasks.  

Using SVM, we experimented with character and word n-gram features weighted by term frequency-inverse term document frequency (tf-idf). We report results for only the most significant ranges, namely, word [1-3] and character [2-7]. As for the classifier training, we used LinearSVC implementation by scikit-learn \footnote{https://scikit-learn.org/}. We use default scikit-learn parameters.

\paragraph{Deep Contextualized Transformer Models (BERT)}
Transformer-based pre-trained contextual embeddings, such as BERT \cite{devlin-etal-2019-bert}, have outperformed other classifiers in many NLP tasks. We used AraBERT \cite{Antoun2020AraBERTTM}, a BERT-based model trained on Arabic news and QARiB \cite{abdelali2021pretraining}, another BERT-model trained on Arabic Wikipedia and Twitter data.
We used ktrain library \cite{maiya2020ktrain} that utilizes Huggingface\footnote{https://huggingface.co/} implementation to fine-tune AraBERT and QARiB. We used learning rate of 8e-5, truncating length of 54 and fine-tuned for 3 epochs. 


\subsection{Results}

\paragraph{Task1: Informativeness} 
For discriminating between more \textit{vs.} less informative tweets,
we designed binary classifiers and reported the results 
in Table \ref{tab:results_all_tasks}.
For baseline, we used majority approach where we assign the label of most frequent class. 
We observed the fine-tuned BERT models, AraBERT and QARiB outperform the SVMs significantly. We noticed AraBERT achieves the highest macro F1 score of 80\%.

\paragraph{Task2: Fine-grained Tweet Categorization}
\revfa{We experiment with fine-grained labels using the multiclass classification setting.}
Due to skewed class distribution, we merged scarce classes (see Table \ref{tab:classes}) and use the hierarchical representation for further classification. For this, we merge \textit{Restrict} and \textit{Request} classes with \textit{Plan}, whereas we integrated \textit{Advice} tweets with \textit{Info-news} due to their similarity in nature. We exclude \textit{Rumor} class since detecting rumors is difficult without any fact-checking or other contextual features. 

We end up with four classes: i) Info-news, ii) Celebrity, iii) Plan, and iv) Less Informative. From Table \ref{tab:results_all_tasks}, we noticed all classifiers outperform the majority baseline. Moreover, we noticed that once again, the fine-tuned BERT models, AraBERT and QARiB outperform the simple SVMs. With F1 score of 67.1, QARiB outperforms AraBERT (F1 score of 64.3) by 2.8\%.
From the confusion matrix (see Figure \ref{fig:conf4}), we observe a confusion for the class \textit{Plan} with \textit{Info\_News}. Such confusion is indeed expected due to the similarity in nature of the tweets. For example, plans introduced by the government are very much similar to the tweets that are discussing the vaccine news or condition to take it.


\begin{table}[!htb]
\scalebox{0.85}{
\centering
\begin{tabular}{llcccc}
\toprule
\textbf{Model} & \textbf{Features} & \textbf{Acc.} &\textbf{P} & \textbf{R} & \textbf{F1}\\ \midrule
\multicolumn{6}{c}{\textbf{Informativeness}} \\ \midrule
Majority & &	79.5 &	39.8	& 50.0 &	44.3\\
SVM	& W[1-3] & 84.0 & 75.7 & 73.1 & 74.3\\
SVM	& C[2-7] &84.9 & 77.6 & 72.9 & 74.8\\
SVM	& C[2-7] + W[1-3] & 84.6 & 76.8 & 73.0 & 74.6\\
QARiB &	& 86.0 & 78.4 & 80 & 79.1\\
AraBERT & & \bf{86.4} &	\bf{78.9} & \bf{81.3} & \bf{80.0}\\ \midrule
\multicolumn{6}{c}{\textbf{Fine-grained categorization (multiclass)}} \\ \midrule
Majority & & 54.4 & 13.6 & 25.0 & 17.6\\
SVM	& W[1-3] & 70.2 & 66.4 & 57.9 & 59.0\\
SVM	& C[2-7] & 71.6 & 66.7 & 58.0 & 58.8\\
SVM	& C[2-7] + W[1-3] & 72.0 & 68.7 & 59.3 & 60.5\\
QARiB & & 72.1 & 66.2 & \bf{68.2} & \bf{67.1}\\
AraBERT	& & \bf{75.4} & \bf{69.2} & 65.1 & 64.3\\\midrule
\multicolumn{6}{c}{\textbf{Stance Detection (multiclass)}} \\ \midrule
Majority & & 81.0 &	27.0 & 33.3	& 29.8\\
SVM	& W[1-3] & 81.6	& 60.8 & 48.6 & 52.1\\
SVM	& C[2-7] & 82.5 & 65.8 & 47.9	& 52.3\\
SVM	& C[2-7] + W[1-3]	& \bf{82.5} & 62.6 & 47.7 & 51.4\\
QARiB & & 81.6 & \bf{64.3} & 62.7 & \bf{63.1}\\
AraBERT & & 82.2 & 61.0	& \bf{65.1} & 62.5\\
\bottomrule
\end{tabular}
}
\caption{Results for different classification tasks.}
\label{tab:results_all_tasks} 
\end{table}

\paragraph{Task3: Stance Detection}
For predicting the stance of the user (tweet), we designed a multiclass  classifier using the aforementioned algorithms. To identify the stance of the tweets, we designed the classifier using 3 classes: positive, negative and neural. 
From our results, in Table \ref{tab:results_all_tasks}, we observe a similar pattern to Task2, where transformers beat SVMs by a significant margin of about 10\%. QARiB achieves the best results with F1 score of 63.1\%. Relatively high error percentage suggests that stance detection is a difficult task for classifiers.
From the per class performance (see Figure \ref{fig:confstance}), we noticed that both neutral and negative stances are confused with the positive ones (the major class).

\section{Error Analysis}
To understand the designed model behaviour, we analyze the errors and confusion made by our best classifier, fine-tuned QARiB for Task2 (fine-grained tweet categorization) and Task3 (stance detection).

\begin{figure}[!h]
\begin{center}
\includegraphics[width=0.70\linewidth]{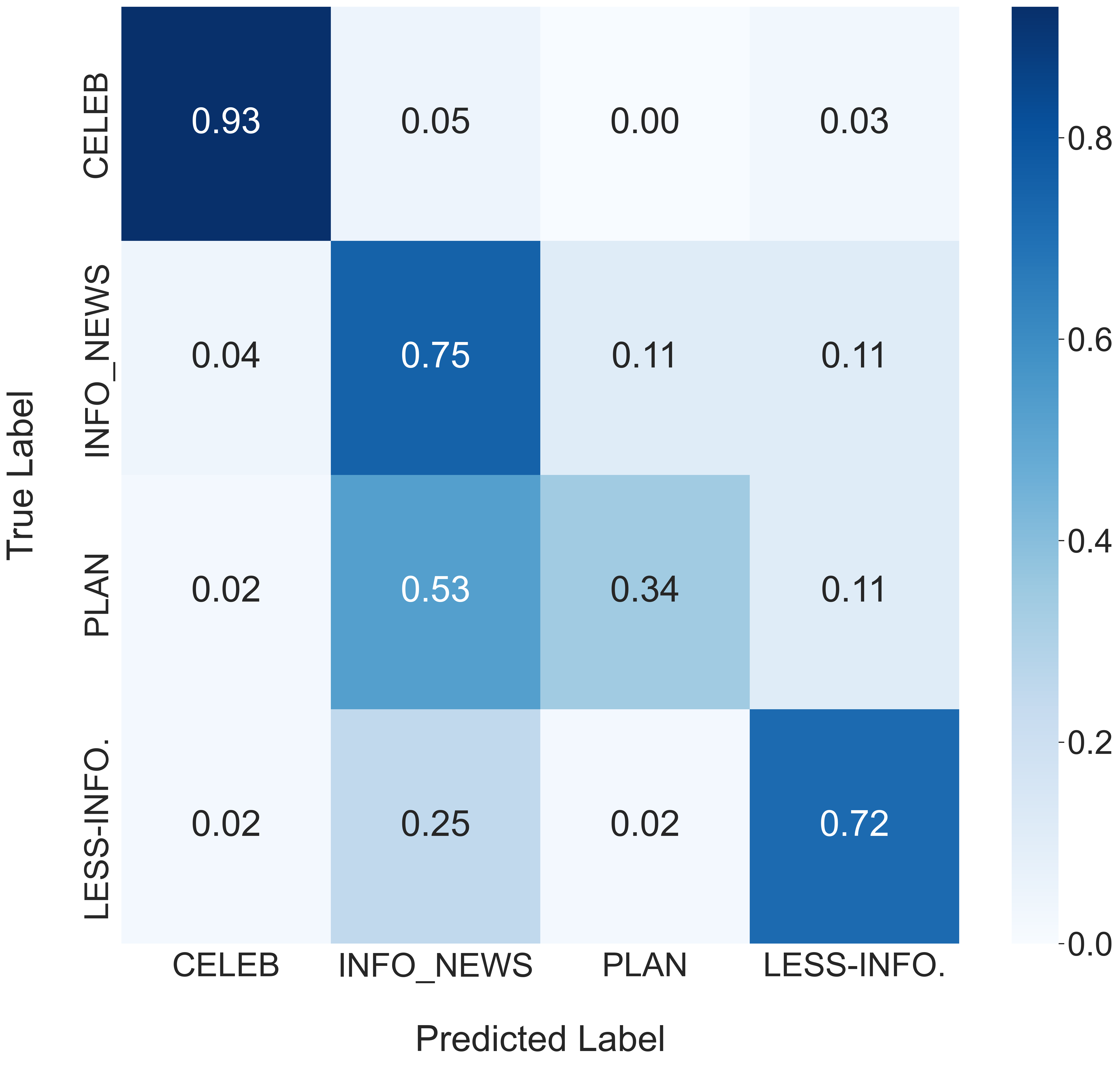} 
\caption{Confusion matrix of fine-grained classification normalized over true labels.}
\label{fig:conf4}
\end{center}
\end{figure}

\begin{figure}[!h]
\begin{center}
\includegraphics[width=0.70\linewidth]{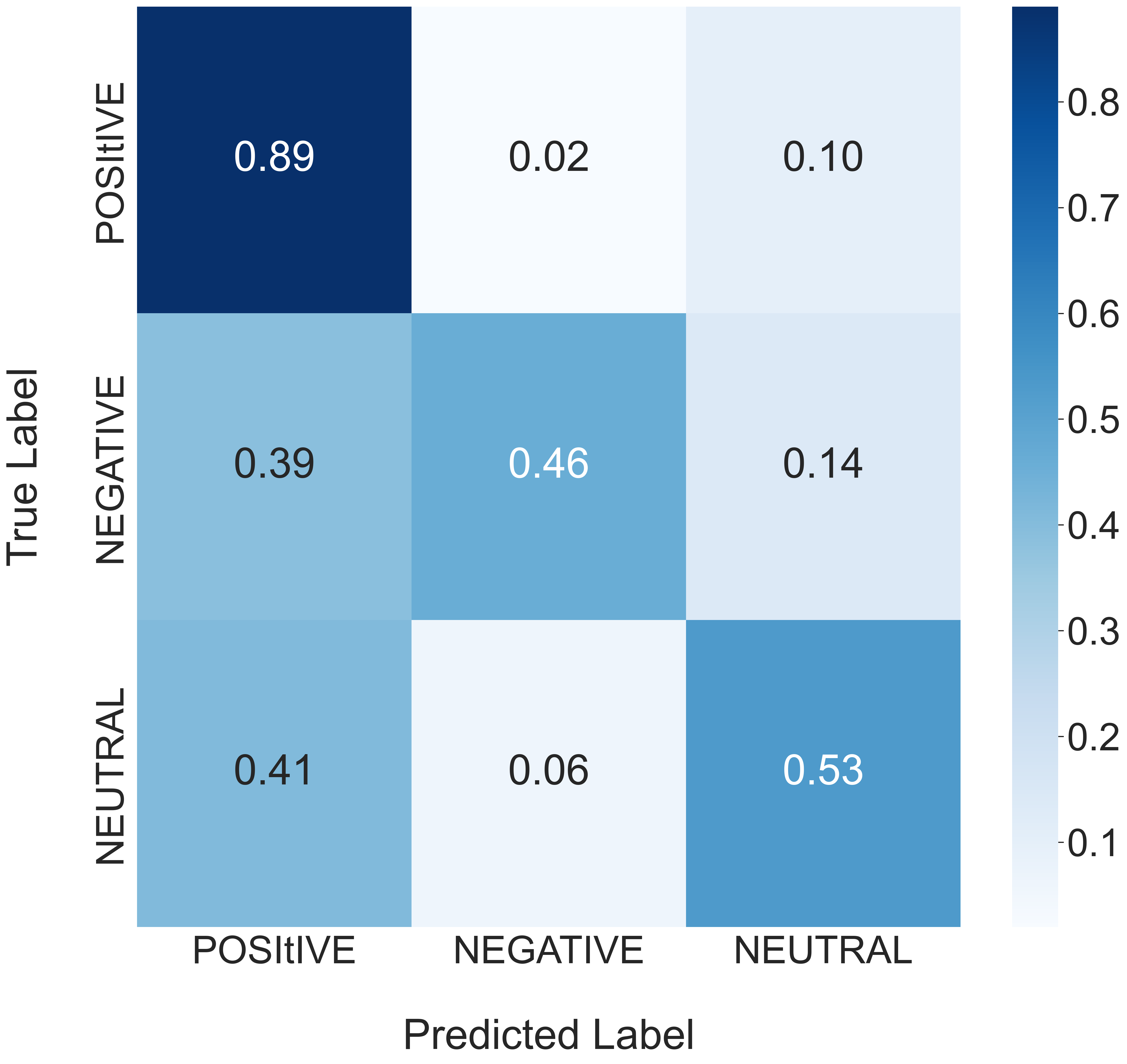} 
\caption{Confusion matrix of stance detection normalized over true labels. 
}
\vspace{-0.3cm}
\label{fig:confstance}
\end{center}
\end{figure}

Figure \ref{fig:conf4} shows the confusion matrix for fine grained classification by QARiB. From the confusion matrix, we can see that most errors stem from \textit{Plan} class misclassified as \textit{Info-news}. Figure \ref{fig:confstance} shows the confusion matrix for stance detection. The confusion matrix shows that most errors stem from Anti-vaccine and Neutral being tagged as Pro-vaccine due to high class imbalance.

\subsection{Classification Errors: Task2}
We picked 200 errors from our best classifier and analyzed them manually. We can summarize most important cases in the following categories:
\begin{itemize}[noitemsep,topsep=0pt,leftmargin=15pt,labelwidth=!,labelsep=.5em]
    \item \textbf{Confusion between 
    classes:} \textit{Info-news} (vaccine) and \textit{Plan} (vaccination process) in the reference or system prediction.
    
    \item \textbf{Annotation errors:} In some cases personal opinions about the vaccine are labeled as informative.
    
    \item \textbf{Multilabel:} Some tweets can have more than one class label. For example, announcement from government about the vaccine followed by details about vaccination plan. We plan to allow multiple labels in the future. We found this case in 10\% of the errors.
    
    \item \textbf{Contextual information:} Need to consider associated multimedia posted with tweet to get the correct prediction. For example, a question about the vaccine and the answer is in an associated video.
\end{itemize}

\subsection{Stance Errors: Task3}
Similarly, for random 200 errors in stance prediction, we found the following issues:
\begin{itemize}[noitemsep,topsep=0pt,leftmargin=15pt,labelwidth=!,labelsep=.5em]
    \item \textbf{Full context:} Need to understand full context including questions and associated multimedia. This includes also considering sarcasm and negation. For example, Is vaccine unsafe? Answer: No.
    
    \item \textbf{Annotation errors:} Labelling a question about taking the vaccine or not as positive stance.
    
    \item \textbf{Ambiguous content:} Errors are due to spam, unrelated or unclear content.
    
    \item \textbf{Mixed/Targeted stance:} For example, refusing vaccines from a certain country but want a safer vaccine.
\end{itemize}

\section{Key Observations}
In this study, 
we show the  popularity of different vaccines,
the common hashtags, e.g., `safe vaccine', present in the data, indicating the main concern of the public towards the vaccine. We also observed different types of rumors spreading the doubts on the safety of vaccination, conspiracy theory and doubts in government assessments and plans. Meanwhile, we also noticed informative tweets confirming vaccine safety, promising fair access and priority and importance of front-liners vaccination. We observed the topics covered in the tweets are mainly health, politics and society centred. From stance timeline, we observed the reliability on the vaccine (pro- stance) increase when leaders/influencers (e.g., kings) takes the vaccine, to set examples.

As for the classification performance, for all the three tasks we noticed transformer architecture outperforms SVMs and present a high performance classifier even with imbalanced class levels. Such performance indicated the efficacy of this data to aid automation of such process.


\section{Conclusion}
We presented and publicly released the first large manually annotated Arabic tweet dataset, \textit{ArCovidVac}, for the COVID-19 vaccination campaign. The dataset of 10k tweets, covering many countries in Arab region, is enriched with different types 
of annotation, including, \textit{(i)} informativeness of the tweets; \textit{(ii)} fine-grained tweet content types with 10 classes; and \textit{(iii)} stance towards vaccination identifying tweets with pro-vaccination (positive), neutral, anti-vaccination content (negative). We performed an in-depth analysis of the dataset considering diverse aspects and presented classification results, which can be used as a benchmark in future studies and aid policymakers in decision making process. 
In the future, we plan to study the dynamics and changes in types/topics of the content and stance towards vaccination in long run.

\section{Bibliographical References}\label{reference}

\bibliographystyle{lrec2022-bib}
\bibliography{lrec2022-example,anthology}


\appendix
\section*{Appendix}
\label{sec:appendix}

\section*{Ethics and Broader Impact}

\subsection*{Dataset Collection}
We collected the dataset using the Twitter API\footnote{\url{http://developer.twitter.com/en/docs}} with keywords that only use terms related to \textit{COVID-19 vaccine}, without other biases. We followed the terms of use outlined by Twitter.\footnote{\url{http://developer.twitter.com/en/developer-terms/agreement-and-policy}} Specifically, we only downloaded public tweets, and we only distribute dehydrated Twitter IDs. We release the dataset by maintaining Twitter data redistribution policy.

\subsection*{Biases}
We note that some of the annotations are subjective. Thus, it is inevitable that there would be biases in our dataset. Yet, we have a very clear instructions, which should reduce biases.

\subsection*{Misuse Potential}
Most datasets compiled from social media present some risk of misuse. We, therefore, ask researchers to be aware that our dataset can be maliciously used to unfairly moderate text (e.g.,~a tweet) that may not be malicious based on biases that may or may not be related to demographics and other information within the text. Intervention with human moderation would be required in order to ensure this does not occur. 

\subsection*{Intended Use}
Our dataset can enable automatic systems for analysis of social media content, which could be of interest to practitioners, social media platforms, and policymakers. Such systems can be used to alleviate the burden for social media moderators, but human supervision would be required for more intricate cases and in order to ensure that the system does not cause harm.

\end{document}